# A General Non-Probabilistic Theory of Inductive Reasoning

*Wolfgang Spohn*
*Institut für Philosophie*
*Universität Regensburg*
*8400 Regensburg*
*West Germany*

## 1. Introduction

Probability theory, epistemically interpreted, provides an excellent, indeed the best available account of inductive reasoning. This is so because there are general and definite rules for the change of subjective probabilities through information or experience; induction and belief change are one and same topic, after all. The most basic of these rules is that one simply has to conditionalize with respect to the information received. Thus, a fundamental reason for the epistemological success of probability theory is that there at all exists a well-behaved concept of conditional probability.

Still, people have, and have reasons for, various concerns over probability theory. One of these is my starting point: Intuitively, we have the notion of plain belief; we believe propositions[1] to be true (or to be false or neither). Probability theory, however, offers no formal counterpart to this notion. Believing $A$ is not the same as having probability 1 for $A$, because probability 1 is incorrigible[2]; but plain belief is clearly corrigible. And believing $A$ is not the same as giving $A$ a probability larger than some $1-\varepsilon$, because believing $A$ and believing $B$ is usually taken to be equivalent to believing $A$-and-$B$.[3] Thus, it seems that the formal representation of plain belief has to take a non-probabilistic route.

Indeed, representing plain belief seems easy enough: simply represent an epistemic state by the set of all propositions believed true in it or, since we assume plain belief to be deductively closed, by the conjunction of all propositions believed true in it. But this does not yet provide a theory of induction, i.e. an answer to the question how epistemic states so represented are changed through information or experience. There is a convincing partial answer: if the new information is compatible with the old epistemic state, then the new epistemic state is simply represented by the conjunction of the new information and the old beliefs. This answer is partial because it does not cover the quite common case where the new information is incompatible with the old beliefs. It is, however, important to complete the answer and to cover this case, too; otherwise, we would not represent plain belief as corrigible. The problem is that there is no good completion. When epistemic states are repre-

---

[1] "Proposition" is the philosophically most common general term for the objects of belief and the one I shall use.

[2] Whatever has probability 1 keeps it, according to all rules of belief change within standard probability theory.

[3] I am here alluding to the so-called lottery paradox, which has gained considerable importance in the writings of H.E. Kyburg, jr., I. Levi, and others. Cf., e.g., the various hints in Bogdan (1982).

315

sented simply by the conjunction of all propositions believed true in it, the answer cannot be completed; and to my knowledge, no other representation of epistemic states has been proposed which would really solve this problem.

In this paper, I want to suggest a solution to this problem. In my (1988), I have more fully argued that this is the only solution, if certain plausible desiderata are to be satisfied. Here, in section 2, I will be content with formally defining and intuitively explaining my proposal. I will compare my proposal with probability theory in section 3. We will see that the theory I am proposing is in important respects structurally homomorphic to probability theory; it thus turns out to be equally easily implementable, but moreover computationally simpler. Section 4 contains a very brief comparison with various kinds of logics, in particular conditional logic, with Shackle's functions of potential surprise and related theories, with Shafer's belief functions, and finally with fuzzy logic.

## 2. Theory

We have first to settle the algebraic framework. Let $W$ be some non-empty set of possibilities (possible worlds, possible courses of events, or what have you). Propositions, denoted by $A, B, C, \ldots$, are represented simply by subsets of $W$. Subfields of the field of all propositions will be denoted by $\mathcal{A}, \mathcal{B}, \mathcal{C}, \ldots$[4] Usually, $W$ will have a structure: there will be a family $(W_i)_{i \in I}$ of variables or factors - where $I$ is some index set and each $W_i$ ($i \in I$) is some non-empty set - such that $W = \prod_{i \in I} W_i$.[5] That is, each $w \in W$ is a function defined on $I$ with $w_i \in W_i$ for all $i \in I$ and thus represents one way how all the variables may get realized. In many physical applications, e.g., each $W_i$ will be identical to the state space and $I$ to the real time axis. For each $J \subseteq I$, $\mathcal{A}_J$ is to be the field $\{A \mid$ for all $w, w' \in W$, if $w_i = w'_i$ for all $i \in J$, then $w \in A$ iff $w' \in A\}$ of all propositions referring at most to the variables in $J$.

The central concept is now easily defined (and afterwards explained):

*Definition 1*: Let $\mathcal{A}$ be a field of propositions. Then $\kappa$ is an $\mathcal{A}$-*measurable natural conditional function* ($\mathcal{A}$-*NCF*) iff $\kappa$ is a function from $W$ into the set $N$ of natural numbers such that $\kappa(w) = 0$ for some $w \in W$ and $\kappa(w) = \kappa(w')$ for all atoms[6] $A$ of $\mathcal{A}$ and all $w, w' \in A$.[7] Moreover, we define for each non-empty $A \in \mathcal{A}$: $\kappa(A) = \min\{\kappa(w) \mid w \in A\}$.[8]

The measurability condition is quite obvious; it requires that an $\mathcal{A}$-NCF does not discriminate possibilities which are not discriminated by the propositions in $\mathcal{A}$.

The crucial question, however, is how to interpret an NCF as an epistemic state. The most accurate answer is to say that an NCF $\kappa$ represents a grading of disbelief: a possibility $w$ with $\kappa(w)=0$ is not disbelieved at all in $\kappa$; if $\kappa(w)=1$, $w$ is disbelieved to degree 1 in $\kappa$; etc. This means that all possibilities $w$ with $\kappa(w)>0$ are believed in $\kappa$ not to obtain; i.e., the true possibility is believed in $\kappa$ to be in $\kappa^{-1}(0) = \{w \mid \kappa(w)=0\}$; and hence the stipulation of Definition 1 that $\kappa^{-1}(0)\neq\emptyset$. A proposition $A$ is believed true in $\kappa$ iff the true possibility is believed in $\kappa$ to be in $A$, i.e. iff $\kappa^{-1}(0)\subseteq A$, i.e. iff $\kappa(-A)>0$.[9] Thus, the set of propositions believed true in $\kappa$ is deductively closed; and it is consistent, because $\kappa^{-1}(0)\neq\emptyset$ - as Definition 1 reasonably guarantees. Note that $\kappa(A)=0$ only means that $A$ is not believed false in $\kappa$; and this is compatible with $\kappa(-A)=0$, i.e. with $A$ also not being believed true in $\kappa$.

---

[4] In the present context we may well assume $W$ to be finite; so, we need not decide which kinds of fields to consider. In the infinite case, complete fields seem to me to be the most appropriate (cf. my (1988); but alternative algebraic frameworks might be used, too.

[5] $\Pi$ denotes the Cartesian product.

[6] $A$ is an atom of $\mathcal{A}$ iff no proper non-empty subset of $A$ ia a member of $\mathcal{A}$.

[7] "Conditional", because these functions can be conditionalized, as we shall see; "natural", because they take natural numbers as values. In my (1988), I have more generally defined "ordinal conditional functions" which take ordinal numbers as values. This generality will not be needed here (all the more as it has some awkward consequences which relate to the fact that addition of ordinal numbers is not commutative).

[8] The latter function for propositions will indeed be the more important one.

We may also talk of integer-valued degrees of firmness of belief, i.e. we may define that $A$ is believed with firmness $m$ in $\kappa$ iff either $\kappa(A)=0$ and $\kappa(-A)=m$ or $\kappa(A)= -m > 0$. Thus, $A$ is believed to be true or false iff, respectively, $A$ is believed with positive or negative firmness.

These explanations well agree with two simple consequences of Definition 1:

*Theorem 1:* Let $\kappa$ be an $\mathcal{A}$-NCF. Then we have:

(1)  for each contingent[10] $A\in \mathcal{A}$, $\kappa(A) = 0$ or $\kappa(-A) = 0$ or both,

(2)  for all non-empty $A,B\in \mathcal{A}$, $\kappa(A\cup B) = \min\{\kappa(A),\kappa(B)\}$.

(1) is the fundamental NCF-law for negation, saying that not both $A$ and $-A$ can be disbelieved. (2) is the fundamental NCF-law for disjunction: It is obvious that $A\cup B$ should be believed at least as firmly as $A$ and $B$. But $A\cup B$ cannot be believed more firmly than both $A$ and $B$; otherwise, it might happen that both $A$ and $B$ are disbelieved, though $A\cup B$ is not. In order to discover a fundamental NCF-law for conjunction, we have to look at conditional NCF-values.

This brings us to the crucial question how epistemic states represented by NCFs are changed through information or experience. Two plausible assumptions provide a complete answer. The first assumption is that, if the information immediately concerns only the proposition $A$, then neither the grading of disbelief within $A$, nor that within $-A$ are changed by that information. We define:

*Definition 2:* Let $\kappa$ be an $\mathcal{A}$-NCF and $A$ a non-empty proposition in $\mathcal{A}$. Then, *the $A$-part of* $\kappa$ is to be that function $\kappa(.|A)$ defined on $A$ for which $\kappa(w|A) = \kappa(w) - \kappa(A)$ for all $w\in A$. If $B\in \mathcal{A}$ and $A\cap B \neq \emptyset$, we also define $\kappa(B|A) = \min\{\kappa(w|A) \mid w\in A\cap B)\} = \kappa(A\cap B) - \kappa(A)$.

The first assumption thus says that an information immediately concerning only $A$ leaves the $A$-part as well as the $-A$-part of $\kappa$ unchanged, i.e. its effect can only be that these two parts are shifted in relation to one another. Definition 2, by the way, already contains the fundamental NCF-law for conjunction:

*Theorem 1* (cont.):

(3)  for all $A,B\in \mathcal{A}$ with $A\cap B \neq \emptyset$, $\kappa(A\cap B) = \kappa(A) + \kappa(B|A)$.

The second assumption is that information about $A$ may come in various degrees

---

[9] $-A$ denotes the complement or the negation of $A$.
[10] $A$ is contingent iff $A$ and $-A$ both are not empty.

of firmness; seeing $A$, e.g., informs about $A$ much more firmly than being told about $A$. Thus, the firmness with which an information is embedded in an epistemic state cannot be fixed once and for all, but has to be conceived as a parameter of the information process itself. In view of the first assumption, this parameter completely determines belief change:

*Definition 3:* Let $\kappa$ be an $\mathcal{A}$-NCF, $A$ a contingent proposition in $\mathcal{A}$, and $m \in N$. Then *the $A,m$-conditionalization $\kappa_{A,m}$ of $\kappa$* is defined as that $\mathcal{A}$-NCF for which $\kappa_{A,m}(w) = \kappa(w|A)$, if $w \in A$, and $\kappa_{A,m}(w) = m + \kappa(w|\text{-}A)$, if $w \in \text{-}A$.

In the $A,m$-conditionalization of $\kappa$, only the $A$-part and the $\text{-}A$-part of $\kappa$ are shifted in relation to one another, and $A$ is believed with firmness $m$, as specified by the conditionalization parameter.

This account of belief change may be generalized. The information may immediately concern not only a single proposition, but a whole field $\mathcal{B}$ of propositions. The parameter characterizing the information process then consists not in a single number, but in a whole $\mathcal{B}$-NCF $\lambda$. And belief change is then defined in the following way:

*Definition 4:* Let $\kappa$ be an $\mathcal{A}$-NCF, $\mathcal{B}$ a subfield of $\mathcal{A}$, and $\lambda$ a $\mathcal{B}$-NCF. Then *the $\lambda$-conditionalization $\kappa_\lambda$ of $\kappa$* is defined as that $\mathcal{A}$-NCF for which for all atoms $B$ of $\mathcal{B}$ and all $w \in B$ $\kappa_\lambda(w) = \lambda(B) + \kappa(w|B)$.

In the $\lambda$-conditionalization of $\kappa$, $\kappa_\lambda(B) = \lambda(B)$ for all $B \in \mathcal{B}$, and only the $B$-parts of $\kappa$, for all atoms $B$ of $\mathcal{B}$, are shifted in relation to one another. Definition 4 corresponds to Jeffrey's generalized conditionalization[11] which is much discussed in probabilistic belief change.

It is to be expected that a workable concept of independence goes hand in hand with this account of conditionalization. This is indeed the case. The following definitions are straightforward:

*Definition 5:* Let $\kappa$ be an $\mathcal{A}$-NCF and $\mathcal{B}$ and $C$ two subfields of $\mathcal{A}$. Then $\mathcal{B}$ and $C$ are *independent with respect to $\kappa$* iff for all non-empty $B \in \mathcal{B}$ and $C \in C$ $\kappa(B \cap C) = \kappa(B) + \kappa(C)$. Furthermore, $\mathcal{B}$ and $C$ are *independent conditional on* the proposition $D$ w.r.t. $\kappa$ iff for all non-empty $B \in \mathcal{B}$ and $C \in C$ $\kappa(B \cap C|D) = \kappa(B|D) + \kappa(C|D)$. If $\mathcal{D}$ is a further subfield of $\mathcal{A}$, then $\mathcal{B}$ and $C$ are *independent conditional on $\mathcal{D}$ w.r.t. $\kappa$* iff $\mathcal{B}$ and $C$ are independent conditional on all atoms $D$ of $\mathcal{D}$ w.r.t. $\kappa$. Finally, these definitions are specialized to two contingent propositions $B$ and $C$ by taking $\mathcal{B}$ as $\{\emptyset, B, \text{-}B, W\}$ and $C$ as $\{\emptyset, C, \text{-}C, W\}$.

How do all the concepts so defined behave? This may not be immediately perspicuous, but the next section will provide a surprisingly powerful answer.

## 3. Comparison with probability theory

The basic definitions and formulae in the previous section look very similar to those in probability theory; we only seem to have replaced the sum, multiplication, and division of probabilities by, respectively, the minimum, addition, and subtraction of NCF-values. In order to see that this is no accident, we have to move for a mo-

---

[11] Discovered by Jeffrey (1965), ch. 11.

ment into the context of non-standard arithmetics and non-standard probability theory:

*Theorem 2:* Let $\mathcal{A}$ be a finite field of propositions. Then, for any non-standard $\mathcal{A}$-NCF[12] $\kappa$ and for any infinitesimal $z$ there is a non-standard probability measure $P$ such that for all $A,B \in \mathcal{A}$ $\kappa(B|A) = n$ iff $P(B|A)$ is of the same order as $z^n$ (i.e. $P(B|A)/z^n$ is finite, but not infinitesimal). In particular we have: whenever $P(C) = P(A) + P(B)$, then $\kappa(C) = \min\{\kappa(A),\kappa(B)\}$; when $P(C) = P(A) P(B)$, then $\kappa(C) = \kappa(A) + \kappa(B)$; $\kappa(B|A) = \kappa(A \cap B) - \kappa(A)$, as desired; and whatever is (conditionally) independent w.r.t. $P$, is so also w.r.t. $\kappa$.

It is thus not surprising that the laws of the concepts introduced in the previous section are simply translations of the laws of the corresponding probabilistic concepts. For instance, the theorem of total probability translates into this (where $A_1,...,A_s$ partition $W$):

(4) $\kappa(B) = \min_{r \leq s} [\kappa(A_r) + \kappa(B|A_r)]$.

Bayes' theorem yields this (with $A_1,...,A_s$ as before):

(5) $\kappa(A_q|B) = \kappa(A_q) + \kappa(B|A_q) - \min_{r \leq s} [\kappa(A_r) + \kappa(B|A_r)]$.

Also, the probabilistic laws of independence and conditional independence hold for NCFs - e.g.:

(6) If $A$ and $C$ are independent w.r.t. $\kappa$, then $B$ and $C$ are independent w.r.t. $\kappa$ iff $A \cup B$ and $C$ are independent w.r.t. $\kappa$ - provided that $A$ and $B$ are disjoint.

Without the proviso, (6) would not necessarily hold. And so on. Let me only mention the most important law concerning conditional independence of subfields. It says in terms of the factorization of $W$ at the beginning of section 2, where $J, K$, and $L$ are pairwise disjoint subsets of the index set $I$:

(7) If $\mathcal{A}_J$ is independent of $\mathcal{A}_K$ conditional on $\mathcal{A}_L$ and independent of $\mathcal{A}_L$ or independent of $\mathcal{A}_L$ conditional on $\mathcal{A}_K$ w.r.t. $\kappa$, then $\mathcal{A}_J$ is independent of $\mathcal{A}_{K \cup L}$ w.r.t. $\kappa$.[13]

These observations have a considerable import. For instance, the theory of probabilistic causation has turned out to be to a large extent a theory of conditional stochastic independence.[14] NCFs would thus allow to extend these ideas to a theory of deterministic causation. In the present context, the crucial observation is, however, that conditional independence is an essential mean for making probability measures computationally manageable. This carries over to the implementation of NCFs. Moreover, all the results and techniques related to such key words as "influence diagram", "Markov field", "causal graph", etc.[15] may be translated into NCF-theory. In particular, Pearl (1986) has shown how probabilistic belief change or updating may be processed in parallel, and Hunter (1988) has given a variant way how to achieve parallel updating of NCFs.

---

[12] This is to mean that $\kappa$ takes non-standard natural numbers as values.

[13] For a proof see my (1988), sect. 6.

[14] Harper, Skyrms (1988), e.g., contains many papers and references supporting this observation.

[15] I mention only Kiiveri et al. (1984) and Pearl (1986). Of course, references could be easily extended.

This does not mean that there are no differences. Certainly, NCF-theory is computationally simpler than probability theory. And when only subjective judgements of experts are to be implemented, it may be easier to elicit these subjective judgements in the coarser terms of NCFs. On the other hand, relative frequencies are so intimately tied to probabilities that I don't see how to reasonably deal with statistical data within an NCF-framework.

## 4. Other comparisons

Though many have proposed representations of epistemic states different from a probabilistic one, I have, to my surprise, nowhere found the simple structure described in section 2. Perhaps the reason is that the importance of stating general and precise rules of belief change, which are tantamount to a theory of induction, has often not been clearly recognized. In any case, this will be my standard criticism of the further comparisons pursued here.

### (a) Various logics

The following idea for modelling belief change has attracted many: Suppose a language with a conditional $\rightarrow$ to be given; represent an epistemic state by a (consistent and deductively closed) set $S$ of sentences of that language; and define the change $S_A$ of $S$ through information $A$ as $S_A = \{B \mid A \rightarrow B \in S\}$.[16] Of course, this idea crucially depends on the properties of $\rightarrow$. E.g., $\rightarrow$ must not be interpreted as material implication. Strict implication will do neither; all the conditionals in the various many-valued logics that have been proposed are unsuited, too[17]; and even the conditionals of the variants of relevance logic turn out to be unhelpful.[18] The best conditional for this purpose is that of conditional logic. Many semantics of conditional logic basically use a well-ordering of possible worlds (or something equivalent or similar).[19] But they don't use numbers and their arithmetical properties. Thus, there arise problems with iterated belief change, and no equally adequate concept of independence is defined within that framework.[20] Finally, epistemic changes as defined in Definition 4 seem completely inaccessible to the whole strategy.

### (b) Plausibility measures

One of the first to propose formal alternatives to the beaten tracks of probability theory was Shackle with his functions of potential surprise.[21] Such a function is a

---

[16] This is the so-called Ramsey test, most thoroughly propounded by Gärdenfors, e.g. in his (1984) and (1986). See also Rott, forthcoming.

[17] As may be easily confirmed with the help of the list in Rescher (1969).

[18] In order to substantiate this remark, we would have to discuss Anderson, Belnap (1975). These remarks are not meant as a criticism, because all the conditionals mentioned were not designed for the present purpose.

[19] Cf., e.g., Lewis (1973) and the overview in Nute (1980).

[20] As is more fully explained in my (1988).

[21] Most extensively presented in Shackle (1969).

function $y$ from the set of propositions into the closed interval $[0,1]$ such that

(8) $y(\emptyset) = 1$,

(9) either $y(A) = 0$ or $y(-A) = 0$ or both,

(10) $y(A \cup B) = \min\{y(A), y(B)\}$.

(9) and (10) are identical with (1) and (2), and (8) arbitrarily fixes the maximal degree of potential surprise to be 1. Thus, Shackle's and my functions only differ in their ranges. This is not a mere technicality, however. There is reason to accept the generalization of (2) or (10) to countable unions (without weakening min to inf), and this forces the range of these functions to be well-ordered. Moreover, I have deliberately avoided a maximal degree of disbelief, because, whenever a proposition acquires this maximal degree, this is incorrigible and cannot be changed any more, according to all rules of belief change. Thus, I object to the possibility accepted by Shackle that propositions different from $\emptyset$ have maximal potential surprise.

The essential point, however, is that Shackle didn't get a grip on conditionalization. This is clear from his proposal

(11) $y(A \cap B) = \max\{y(A), y(B|A)\}$,

where he left $y(B|A)$ in fact undefined.

Similar remarks apply to the plausibility indexing proposed by Rescher, e.g. in his (1976), and to the inductive probability of Cohen (1977) (which is not mathematical probability).

*(c) Shafer*

Shafer (1976), p.224, shows that Shackle's theory is a special case of his: the function $y$ is a degree of doubt derived from a consonant belief function in the sense of Shafer iff it satisfies (8)-(10). Since Dempster's rule of combination governs belief change for Shafer's belief functions in general, we may expect it to complete Shackle's theory. It indeed does, but in a different way than I did in section 2:

According to Shafer (1976), pp.43+66f., there are also conditional degrees of doubt given by the formulae

(12) $y(B|A) = [y(A \cap B) - y(A)] / [1 - y(A)]$.

Apart from the denominator, this looks like my Definition 2. However, $y(.|A)$ here represents the degree of doubt which results from combining the old belief function with the belief function $Bel$ defined by: $Bel(B) = 1$, if $A \subseteq B$, and $Bel(B) = 0$ otherwise; and this function makes $A$ incorrigibly certain, according to Shafer's theory. Thus, we should rather know how Shafer processes evidence which makes $A$ less than incorrigibly certain, since this is what the above Definition 3 accomplishes. Shafer does this by combining the old belief function with some belief function $Bel_s$ defined by: $Bel_s(B) = 1$, if $B=W$, $Bel_s(B) = s$, if $A \subseteq B \neq W$, and $Bel_s(B) = 0$ otherwise ($0<s<1$). (In a sense, $s$ corresponds to the $m$ of Definition 3.) But now the problem arises that, if the old belief function is consonant, its combination with $Bel_s$ will in general not be consonant. Thus, my dynamics of belief moves within the set of all NCFs, whereas the set of all consonant belief functions in Shafer's sense is not closed with respect to Shafer's dynamics. This basic difference entails that Shafer's theory is not a generalization of the NCF-theory presented here, but is rather in conflict with it.

*(d) Fuzzy logic*

The NCF-formalism seems quite incomparable to fuzzy logic. This is more perspicuous when we notice that Zadeh's preferred base logic for his fuzzy logic is Lukasiewicz' logic with infinitely many truth-values[22], which, as noted in (a), is quite different from NCF-theory. Indeed, both theories also seem intuitively incomparable. Fuzzy logic deals with vague expressions and approximate reasoning. But inductive reasoning as such need in no way be vague or approximate; nor is reasoning inductive simply because it is approximate. Thus, I think that the two theories just have different intended applications. This, however, does not rule out the possibility that there is a reasonable and useful fuzzification of NCF-theory.[23]

---

[22] Cf. e.g. Zadeh (1975).

[23] I am greatly indebted to Dan Hunter for developing my philosophical stuff for AI and for encouraging me to present it there.